\documentclass[5pt]{article}

\usepackage[letterpaper]{geometry}
\usepackage{spconf,amsmath,epsfig}
\usepackage{enumitem}

\usepackage{fancyhdr}
\usepackage{spconf,amsmath,epsfig}
\usepackage{amssymb}
\usepackage{graphicx}
\usepackage{enumerate}
\usepackage{multirow} 
\usepackage{subfig} 
\usepackage{float}    
\usepackage{url}
\usepackage{color}
\usepackage[colorlinks,urlcolor=black]{hyperref}
\usepackage{diagbox}
\usepackage[ruled, linesnumbered]{algorithm2e}
\usepackage{algpseudocode}
\usepackage{tabularx,booktabs}
\newcolumntype{Y}{>{\centering\arraybackslash}X}

\usepackage{setspace}
\usepackage{fancyhdr}
\usepackage{leftidx}
\usepackage{bmpsize}
\usepackage{cite}
\usepackage{listings}
\usepackage{makecell}

\begin{document}\sloppy

\def\x{{\mathbf x}}
\def\L{{\cal L}}

\title{Occluded Person Re-identification}
%
\name{Jiaxuan Zhuo$^{1,3,4}$, Zeyu Chen$^{1,3,4}$, Jianhuang Lai$^{1,2,3}$\textsuperscript{*}\thanks{\textsuperscript{*}Corresponding author}, Guangcong Wang$^{1,3,4}$}
\address{$^1$Sun Yat-Sen University, Guangzhou, P.R. China\\
  $^2$XinHua College, Sun Yat-sen University, Guangzhou, P.R. China\\
  $^3$Guangdong Key Laboratory of Information Security Technology\\
  $^4$Key Laboratory of Machine Intelligence and Advanced Computing, Ministry of Education\\
  \{zhuojx5,chenzy5,wanggc3\}@mail2.sysu.edu.cn, stsljh@mail.sysu.edu.cn}

%
%

\maketitle

\begin{abstract}
Person re-identification (re-id) suffers from a serious occlusion problem when applied to crowded public places. In this paper, we propose to retrieve a full-body person image by using a person image with occlusions. This differs significantly from the conventional person re-id problem where it is assumed that person images are detected without any occlusion. We thus call this new problem the occluded person re-identitification. To address this new problem, we propose a novel Attention Framework of Person Body (AFPB) based on deep learning, consisting of 1) an Occlusion Simulator (OS) which automatically generates artificial occlusions for full-body person images, and 2) multi-task losses that force the neural network not only to discriminate a person's identity but also to determine whether a sample is from the occluded data distribution or the full-body data distribution. Experiments on a new occluded person re-id dataset and three existing benchmarks modified to include full-body person images and occluded person images show the superiority of the proposed method.
\end{abstract}
\section{Introduction}
\label{sec:intro}

Person re-identification (re-id) aims to re-identify a target person across multiple non-overlapped cameras, which has been applied to enhance the security in many important public spaces, especially crowded ones, e.g., airports, railway stations, malls and hospitals. However, when conducting person re-id in these crowded places, occlusion is an unavoidable problem. For example, a person/criminal may be occluded by other persons in the scene, or static obstacles such as cars, pillars, walls, etc. Considering the significance of the occlusion problem, it is essential to seek an effective method to search full-body person images given a person image with occlusions as a probe (Fig. 1). We call this the occluded person re-identification problem.

\begin{figure}[!t]
\centering
  \includegraphics[width=0.45\textwidth]{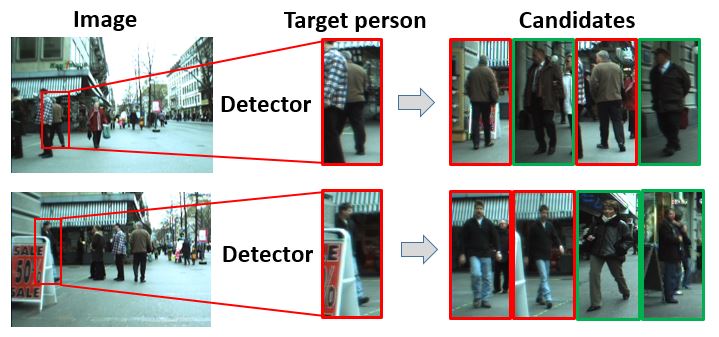}    
  \caption{Illustration of occluded person re-id. Left: the original images in video surveillance; Middle: occluded target persons detected by existing pedestrian detector; Right: our goal is to retrieve a full-body person given a person with occlusions. Red bounding box means the same one, while the green means different ones.}
  \vspace{-0.3cm}  
\end{figure}

There are three realistic challenges becoming the bottleneck for solving the occluded person re-id. First, occlusions lead to not only the loss of target information but also the interference of occluded information. Occlusions with diverse characteristics, such as colors, sizes, shapes and positions, deteriorate global representations for the person re-id. So it is hard to learn a robust feature representation for occluded persons. Second, one may resort to local/part-based representations for the occluded person images. An intuitive method is to detect non-occluded body parts using body part detectors and then match the corresponding body parts in the gallery. However, extra annotations are needed for the body detector learning. Even worse, sometimes occluded body parts are the key discriminative parts while non-occluded body parts share a similar appearance. Third, since most existing methods implicitly assumed that the appearance of full body for a person is readily available while a person image with occlusions is an invalid sample, there are few public datasets for the occluded person re-id to learn a suitable model, especially for deep learning.
\begin{figure}[!t]
\centering
  \includegraphics[width=0.5\textwidth]{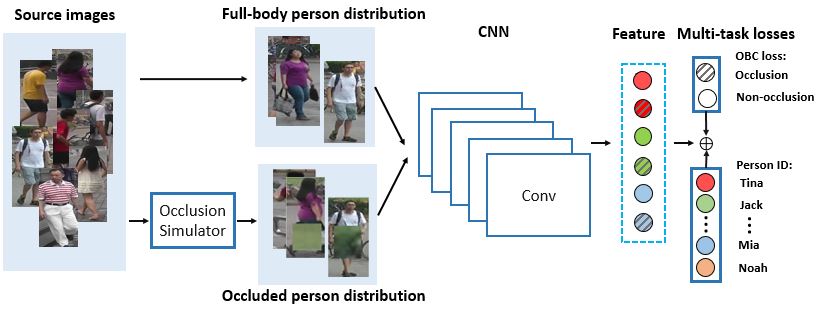}    
  \caption{Overview of our approach. Given source images, we first use an OS to generate artificial occluded data and then jointly encode the source data and occluded data to a CNN network with multi-task losses.}
  \vspace{-0.3cm}  
\end{figure}

To solve this challenging problem, we propose a novel deep learning framework for the occluded person re-id, called Attention Framework of Person Body (AFPB) (Fig. 2). Specially, the AFPB consists of two components. First, an Occlusion Simulator (OS) is used to automatically generate plenty of artificial occluded person images by randomly adding background patches to full-body person images. The artificial occluded person images are thus formed as an artificial occlusion set. The artificial occlusion set and the source (full-body) set are then jointly used to learn a robust feature representation for the occluded person re-id. Second, multi-task losses, i.e., identification loss and occluded/non-occluded binary classification (OBC) loss, are integrated into the AFPB framework. Surprisingly, the simple OBC loss brings an impressive improvement by determining whether a sample is from artificial occlusion set or source images.

The AFPB method can be formulated as an ``attention" model by resorting to the OS module. Different from the conventional attention scheme, the AFPB framework gradually pays attention to the person body by watching different kinds of occluded persons that generated by the OS module. The AFPB framework can be also implicitly explained as an encoder of prior knowledge that allows one to integrate extra expert knowledge into the deep framework. Specially, the OS module of the AFPB integrates the occlusion information by generating lots of artificial knowledge-related samples towards mimicking the real world, while the OBC loss aims to determine whether a sample is from the occluded person set or the full-body person set, so as to encode prior information into the framework. In addition, the identification loss can discriminate the person identity given a person image no matter whether the person is occluded or not. This constraint also forces the framework to focus on person body parts.

The AFPB method can be trained end-to-end by SGD with backpropagation, and can be easily implemented using common libraries (e.g., Caffe). In summary, this paper makes three main contributions.

\begin{itemize}
  \item It is the first attempt to define the occluded person re-id problem which commonly occurs in realistic scenes and applications.
  \item For the occluded person re-id task, we propose an Attention Framework of Person Body (AFPB) which gradually pays attention to the person body by watching different kinds of occluded persons. An Occlusion Simulator (OS) is used to generate artificial samples to encode prior knowledge while multi-task losses are used to learn a robust feature representation against the occluded problem.
  \item We create a new dataset called Occluded-REID and two modified datasets called P-DukeMTMC-reID and P-ETHZ that are sorted from DukeMTMC-reID \cite{zheng2017unlabeled} and ETHZ \cite{ess2008mobile} respectively for the occluded person re-id.
\end{itemize}

\section{RELATED WORK}
Typical person re-id works mainly consist of two steps: feature extraction and metric learning. The first step aims to extract a robust and distinctive feature representation which is invariant to the challenges such as illumination, viewpoint, and occlusion, etc. The second step learns metrics or subspaces for better matching such that distances of the same class are closer than those of the different ones.

Recently, with the development of deep learning \cite{wang2017deep}, there are three kinds of network frameworks applied in person re-id, i.e., classification networks, siamese networks and triplet networks. Classification networks regard the person re-id problem as a classification problem and directly extract discriminative features due to the superior performance of Convolutional Neural Networks (CNN) on large-scale datasets. For example, Xiao et al.\cite{xiao2016learning} jointly trained a classification model on multiple domains by using a Domain Guided Dropout (DGD) method to improve the performance. Siamese networks take image pairs as input and compute the similarity using a contrastive loss. For example, Ahmed et al. \cite{ahmed2015improved} computed the neighborhood differences of an image pair to learn a metric indicating whether the two input images depict the same person. Triplet networks, in virtue of similarity among three input images, is an extension version of siamese networks. Based on \cite{ding2015deep}, Wang et al. \cite{wang2017p2snet} developed a point-to-set triplet for the image-to-video person re-id.

Although person re-id methods have been studied excellently \cite{wang2017p2snet, xiao2016learning, ahmed2015improved, ding2015deep, chen2018person, chen2016deep, shi2015person, guo2014multi}, few works make initial attempts to solve the occluded person re-id problem. Among the works that share similar idea to ours, partial person re-id \cite{zheng2015partial} aims to match the probe partial image with the gallery full-body image, which is towards providing a picture to the occluded problem in person re-id. However, they only focused on the matching between non-occluded body parts and full-body parts. Some critical problems still need to be solved:
1) the output of existing pedestrian detectors for an occluded person image often includes a part of person body together with other occlusions instead of only a partial body. That is, occlusions have to be removed by a manual cropping operation in \cite{zheng2015partial}, which is unrealistic in practice;
2) the patch-based matching method proposed in \cite{zheng2015partial} needs a large account of calculations without considering an attention scheme.
Our work based on a CNN network differs from \cite{zheng2015partial}, as we directly compute the matching between occluded person images and full-body person images and propose an AFPB framework that automatically focuses on the person body by watching various occluded person data generated by an OS.

\section{METHODOLOGY}

We propose a deep learning framework, Attention Framework of Person Body (AFPB), as shown in Fig. 2. The AFPB includes two main components: Occlusion Simulator (OS) and multi-task losses. The OS aims to generate artificial occluded person data which is used to simulate a variety of occluded cases using source (full-body person) data. Next, full-body person data and occluded person data are jointly trained on the CNN network with multi-task losses, i.e., identification loss and occluded/non-occluded binary classification (OBC) loss. In general, the AFPB forces the feature representation to pay more attention to person body parts by encoding prior information of occlusion into the framework. In this section, we detail these stages and visualize of the attention results.

\subsection{Occlusion Simulator (OS)}
As is mentioned in Section 1, with the limitation of inadequate occluded person re-id data, it is hard to train a suitable deep model for the occluded person re-id. One would think whether plenty of occluded person re-id data can be created based on existing data. A good idea to this problem is to automatically generate artificial occluded person data from the full-body person data. In this way, not only can it simulate a variety of occluded cases but can bring diversified information to the whole system. Based on the above assumptions, we design an Occlusion Simulator (OS) to generate artificial occluded person images. Next, we jointly train a CNN network on the source/full-body person data and the occluded person data. The network would pay more attention to person body parts by watching various occluded person images in person re-id. The specific implementation is as follow.

Suppose we have an original full-body data set X, which consists of N images of M identities. Let ${\{(x_{i}^{(j)}, y_{i})_{j=1}^N\}_{i=1}^M}$ denote all the samples in $X$, where $x_{i}^{(j)}$ is the $j_{th}$ image of the $i_{th}$ person and $y_{i}\in\{1, 2, \ldots, M\}$ is the identity. The Occlusion Simulator can be formulated as an image-image mapping function F : $X \rightarrow Z$, where Z is an artificial occluded person data set generated from the real images set X. The mapping F is achieved by a simple but effictive way where a random patch from the background of source images is used as an occclusion to cover a part of person body. Let ${\{(z_{i}^{(j)}, y_{i})_{j=1}^{N} \}_{i=1}^M}$ denote all the samples in $Z$, where $z_{i}^{(j)}$ is the $i_{th}$ image of the $j_{th}$ person generated from $x_{i}^{(j)}$. We finally merge X and Z into a combined set. The procedure is shown in Algorithm 1.

\begin{algorithm}
  \caption{Occlusion Simulation}
  \SetKwInOut{Input}{input}
  \Input{A full-body set $X$ (with $N$ real images)\\
    Size of background patch $s$\\
    Range of occluded area ratio $\left[r_{1}, r_{2} \right ]$
  }
  \KwOut{Artificial occlusion set $Z$, Combined set O}
  \While{$N > 0$}{
    Randomly select an image $I$ with label $y$ from $X$\\
    Select one occluded area $S_{o} \in \left[S_{I} \times r_{1}, S_{I} \times r_{2} \right ]$\\  
        $S_{b} = s \times s\leftarrow$ crop a background patch from $I$\\
        $S_{o}\leftarrow$ resize $S_{b}$ \\
        $I^{*}\leftarrow$ a random position of $I$ is covered by $S_{o}$\\
        Put $I^{*}$ with label $y$ to set $Z$ \\
        $N\leftarrow N-1$\\
  }
  Combined set $O\leftarrow$combine set $X$ and set $Z$
\end{algorithm}

We aim to learn a generic feature extractor $h(\cdot)$ which makes descriptors of the same person closer while those of different ones more distinct. In our framework, we train a CNN with identification loss to recognize the identity of each person. When only training on the full-body set X, the objective function is
\begin{equation}
\underset{f,h}{\arg\min}\sum_{i=1}^M\sum_{j=1}^NL^P(f(h(x_{i}^{(j)})), y_i) \\
\end{equation}
where $f(\cdot)$ is the identification classifier in person re-id and $L^P(\cdot)$ is the identification loss function. After generating artificial occlusion set Z from X, we combine two sets together, so the objective function is given by
\begin{equation}
\underset{f,h}{\arg\min}\sum_{i=1}^M\sum_{j=1}^NL^P(f(h(x_{i}^{(j)})), y_i)+\sum_{i=1}^M\sum_{j=1}^{N}L^P(f(h(z_{i}^{(j)})) ,y_i) \\
\end{equation}
It can be seen that the objective function makes both $h(x_i)$ and $h(z_i)$ closer to $y_i$, so it would force $h(z_i)$ to be more similar to $h(x_i)$. It is intuitively explained that the network has learned how to pay more attention to the key person body parts rather than occlusions or backgrounds by watching lots of occluded person data and source data.

\subsection{Multi-task losses}

Along with the identification loss, an occluded/non-occluded binary classification (OBC) is used to determine whether a sample is from an occluded person distribution or a full-body person distribution, such that our framework can identify the person on the basis of discriminating whether a person body is occluded or not. When integrating two losses into a unified framework, the AFPB method is further learned to extract a robust and discriminative feature representation for the occluded person re-id. In this way, the OBC loss encodes prior information of occlusion into the framework.


We treat the person re-id as a classification problem and use the softmax loss as the identification loss. Suppose the original full-body set has K identities, a K-class softmax loss of person re-id classifier is given by
\begin{equation}
L^P(\hat{y_i}, y_i) = \sum_{k=1}^K\{y_i=k\}log\frac{e^{\hat{y_i}}}{\sum_{k=1}^Ke^{\hat{y_k}}} \\
\end{equation}
where $\hat{y_i}$ is the prediction score in person re-id classifier of the $i_{th}$ training sample for the $k_{th}$ class. As such, the OBC loss is given by
\begin{equation}
L^O({\hat{y_i}}', {y_i}') = \sum_{c=0}^C\{{y_i}'=c\}log\frac{e^{{\hat{y_i}}'}}{\sum_{c=0}^Ce^{{\hat{y_c}}'}} \\
\end{equation}
where $\hat{y_i}^{'}$ is the prediction score in occlusion classifier of the $i_{th}$ training sample, and $c \in \{0, 1\}$ where $c = 0$ indicates occluded persons and $c = 1$ otherwise. Combining the identification loss and OBC loss, the multi-task losses are formulated as
\begin{equation}
L = \alpha L^P(\hat{y_i}, y_i) + (1-\alpha) L^O({\hat{y_i}}', {y_i}') \\
\end{equation}
where $\alpha \in (0,1)$ is a hyperparameter which balances two respective losses. Generally, it is reasonable to set $\alpha \geq 0.5$ because $L^P(\hat{y_i}, y_i)$ is the protagonist and $L^O({\hat{y_i}}', {y_i}')$ is an assistance for $L^P(\hat{y_i}, y_i)$. With multi-task losses, CNN network has the discriminability to identify the person no matter whether a person is occluded or not. That is, if a full-body person image is availuable, the network can exploit the entire person structure information. If a person is occluded, the network can focus on the key body parts.

Through this process, our framework can focus on person body parts to learn a robust feature representation against occlusions in the real world. Fig. 3 shows saliency maps generated by average pooling all feature maps of the last convolution layer. It proves that our framework can pay more attention to person body parts rather than occlusions or backgrounds.
\begin{figure}[!tp]
\centering
  \includegraphics[width=0.46\textwidth,height=0.23\textheight]{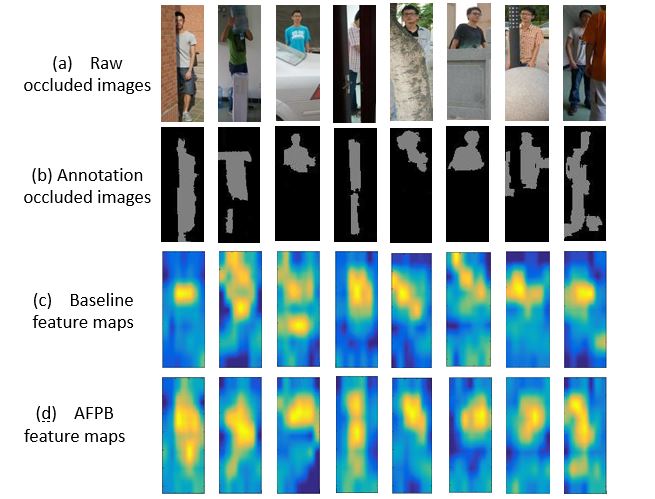}    
  \caption{Examples of (a) raw occluded images, (b) occluded images with manual annotations of person body parts, (c) saliency maps of the baseline network, (d) saliency maps of the AFPB. The salient region reveals the part that the network representation focuses on.}
  \vspace{-0.3cm} 
\end{figure}

\section{EXPERIMENTS}

\subsection{Setup}
\noindent\textbf{Datasets.} We evaluate the proposed method on four datasets: Occluded-REID, Partial-REID, P-DukeMTMC-reID and P-ETHZ, each of which is organized into two parts: occluded person images and full-body person images (see Fig. 4).

\noindent\textbf{Occluded-REID dataset} is a new dataset captured by mobile camera equipments, which consists of 2000 images of 200 occluded persons. Each identity has 5 full-body person images and 5 occluded person images with different types of severe occlusions. All images with different viewpoints and backgrounds are resized to 128 $\times$ 64. This dataset will be released later.

\noindent\textbf{P-DukeMTMC-reID dataset and P-ETHZ dataset} are modified from DukeMTMC-reID dataset\cite{zheng2017unlabeled} and ETHZ dataset\cite{ess2008mobile}. They contain images with target persons occluded by different types of occlusion in public, e.g., people, luggages, cars and guideboards. We select identities with both full-body person images and occluded person images. After the arrangement, there are 24143 images of 1299 IDs in P-DukeMTMC-reID and 3897 images of 85 IDs in P-ETHZ, respectively. Both datasets will also be released later.

\noindent\textbf{Partial-REID dataset} is the first dataset for partial person re-id\cite{zheng2015partial}, which includes 900 images of 60 persons, with 5 full-body person images, 5 partial person images and 5 occluded person images each identity. The images were collected at a university campus with various viewpoints and occlusions.


\noindent\textbf{Experimental setting.} We take occluded person images as the probes, full-body person images as the galleries and randomly select half of the identities for training and the rest for test. We report the results trained on a baseline network, ResNet-50\cite{he2016deep}. Both single-shot (N=1) and multi-shot (N=2, 3, 4, 5) experiments were conducted with the initial learning rate of 1e-3, $batch\:size = 20$ and $\alpha = 0.8$ for 50K iterations.

\noindent\textbf{Evaluation metric.} In matching produce, we calculate the similarities between each probe and all the gallery images by $L_2$ distance. The widely-used Cumulation Matching Characteristics (CMC) curve\cite{wang2017p2snet} and rank-1 rate are used for quantitative evaluations of person re-id task. The experiments are repeated 10 times to gain the average results.

\noindent\textbf{Data augmentation.} In our experiment, we resize all images into 240 $\times$ 240 and crop a center region of 224 $\times$ 224 with a small random perturbation to augment the training data.

\begin{figure}[tp]  
  \centering
  \includegraphics[width=0.45\textwidth]{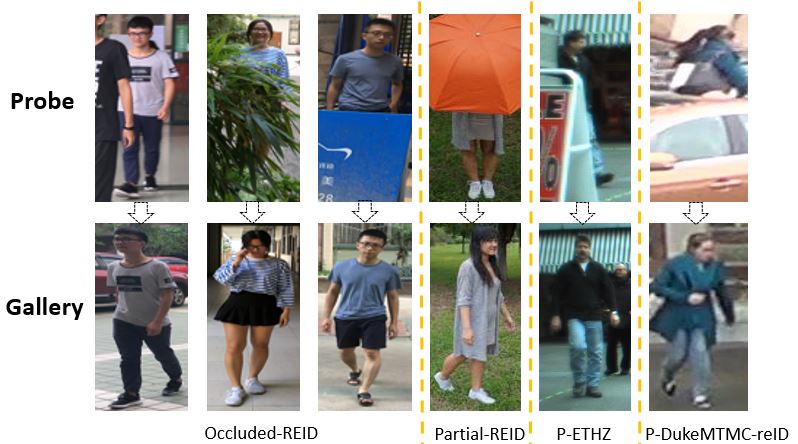}    
  \caption{Examples of datasets. Upper: occluded person images; Below: full-body person images.}
  \vspace{-0.5cm}  
\end{figure}

\subsection{Performance of AFPB framework}
To evaluate the performance of our Attention Framework of Person Body (AFPB), we compare it with three networks: 1) the baseline network, ResNet-50 pretained on a large-scale person re-id dataset MARS \cite{zheng2016mars}, 2) the baseline network with the first component of the AFPB, Occlusion Simulator (OS), 3) the baseline network with the second component of the AFPB, multi-task losses, on four datasets, Occluded-REID, Partial-REID, P-DukeMTMC-reID and P-ETHZ. As is shown in Table 1 and Fig. 5, our AFPB framework significantly outperforms the baseline by a large margin (improve rank-1 accuracy by $10.9\%$, $8.7\%$, $7.0\%$ and $7.1\%$, respectively), which shows the effectiveness of our framework. Besides, the performance of ResNet-50 with the OS and multi-task losses are also better than that of the baseline but worse than that of our AFPB framework. It illustrates that both components of the framework make contribution to the proposed framework and it could achieve more excellent performance by combining them together because two components of the AFPB form a complementary relationship to deal with various occluded situations.

Besides, we evaluate occlusion awareness and attention performance of AFPB in two experiments. First we test the accuracy of OBC on  the trained model. The classification accuracies are 88.50\%, 85.33\%, 91.75\% and 73.88\% on Occluded-REID, Partial-REID, P-DukeMTMC-reID and P-ETHZ, which demonstrates the OBC loss offers occlusion awareness to the AFPB. Then we compute the detection precision\cite{borji2015salient} which is the ratio of salient regions in saliency maps to our manual annotations of body parts on Occluded-REID and Partial-REID (Fig. 3). Table 2 shows our method exceeds the baseline network on the detection precision by 6.21\% and 7.13\%, respectively, which confirms the AFPB has the superiority to focus on the person body.

\subsection{Comparison with the state-of-the-art}
We compare our method with the state-of-the-art methods on the Occluded-REID, Partial-REID, P-DukeMTMC and P-ETHZ in Table 3 and Fig. 6. We collect seven methods including (A) four methods of hand-crafted features and distance metrics and (B) three methods based on deep learning. It is evident that our method presents the best performance in all categories generally and surpasses the $2^{nd}$ best in rank-1 by 2.35\%, 2.33\%, 0.97\% and 3.72\% on Occluded-REID, Partial-REID, P-DukeMTMC-reID and P-ETHZ respectively. Generally the performances of methods in (B) are better than those in (A) due to the powerful learning ability of deep nerual network to automatically learn and update the model. Some methods in (B) \cite{xiao2016learning} do not show good performance beacause their models concern about data with full-body images regardless of the occluded data domain. Differently, our methods can learn to determine whether a person is occluded or not and extract robust feature representation based on prior knowledge of occlusion. In this way, our method shows the superiority in the occluded person re-id.

\begin{table}[!tp]
  \caption{Performance of AFPB in single/multi-shot} \label{tab:cap}
  \renewcommand\arraystretch{1.0} 
  \tabcolsep0.4pt 
  \scriptsize  
  \doublerulesep=0.4pt

\begin{tabular}{|p{0.15cm}|c|c|c|c|c|c|c|c|c|c|}
    \hline
\multirow{2}{*}{\rotatebox{90}{\tiny DATA}}
    &\multirow{2}{*}{\rotatebox{0}{METHODS}} & \multicolumn{3}{c|}{N=1} & \multicolumn{3}{c|}{N=2} & \multicolumn{3}{c|}{N=5}\\     
    \cline{3-11}  
    &  & r=1 & r=5 & r=10 & r=1 & r=5 & r=10 & r=1 & r=5 & r=10 \\  
    \hline
\multirow{4}{*}{\rotatebox{90}{\tiny Occluded-REID}}    
    & ResNet-50\cite{he2016deep} & 57.27 & 81.40 & 89.66 & 76.37 & 94.43 & 97.96 & 90.00 & \textbf{\textcolor{red}{99.00}} & 99.00 \\
    & ResNet-50+Occlusion Simulator & 63.95 & 85.62 &  92.05 & 82.82 & 96.34 & 98.71 & 90.00 & \textbf{\textcolor{red}{99.00}} & \textbf{\textcolor{red}{100.00}} \\
    & ResNet-50+Multi-task Losses & 59.13 & 82.39 & 90.42 & 78.03 & 94.72 & 98.23 & 86.00 & \textbf{\textcolor{red}{99.00}} & \textbf{\textcolor{red}{100.00}} \\
    & ResNet-50+Ours & \textbf{\textcolor{red}{68.14}} & \textbf{\textcolor{red}{88.29}} & \textbf{\textcolor{red}{93.67}} &  \textbf{\textcolor{red}{86.51}} & \textbf{\textcolor{red}{97.26}} & \textbf{\textcolor{red}{98.88}} & \textbf{\textcolor{red}{94.00}} & \textbf{\textcolor{red}{99.00}} & \textbf{\textcolor{red}{100.00}} \\
  \hline  
\multirow{4}{*}{\rotatebox{90}{\tiny Partial-REID}}       
    & ResNet-50\cite{he2016deep} & 69.82 & 89.68 & 96.10 & 84.53 & 96.45 & 98.98 & 90.00 & \textbf{\textcolor{red}{100.00}} & \textbf{\textcolor{red}{100.00}} \\
    & ResNet-50+Occlusion Simulator & 74.29 & 92.91 & 97.69 & 89.63 & 98.89 & 99.69 & 93.33 & \textbf{\textcolor{red}{100.00}} & \textbf{\textcolor{red}{100.00}} \\
    & ResNet-50+Multi-task Losses & 71.26 & 89.77 & 96.37 & 85.89 & 96.58 & 99.00 & 90.00 & \textbf{\textcolor{red}{100.00}} & \textbf{\textcolor{red}{100.00}} \\
    & ResNet-50+Ours & \textbf{\textcolor{red}{78.52}} & \textbf{\textcolor{red}{94.87}} & \textbf{\textcolor{red}{98.03}} & \textbf{\textcolor{red}{91.25}} & \textbf{\textcolor{red}{99.20}} & \textbf{\textcolor{red}{99.77}} & \textbf{\textcolor{red}{96.67}} & \textbf{\textcolor{red}{100.00}} & \textbf{\textcolor{red}{100.00}} \\
\hline
\multirow{4}{*}{\rotatebox{90}{\tiny P-DukeMTMC}}   
    & ResNet-50\cite{he2016deep} & 39.16 & 57.13 & 64.35 & 44.62 & 63.58 & 70.73 & - & - & - \\
    & ResNet-50+Occlusion Simulator & 40.50 & 57.73 & 65.22 & 46.67 & 65.27 & 72.79 & - & - & - \\
    & ResNet-50+Multi-task Losses & 39.04 & 57.21 & 64.80 & 45.24 & 64.37 &  71.66 & - & - & - \\
    & ResNet-50+Ours & \textbf{\textcolor{red}{46.15}} & \textbf{\textcolor{red}{63.47}} & \textbf{\textcolor{red}{70.67}} & \textbf{\textcolor{red}{51.95}} & \textbf{\textcolor{red}{70.40}} & \textbf{\textcolor{red}{77.22}} & - & - & - \\   
    \hline
\multirow{4}{*}{\rotatebox{90}{\tiny P-ETHZ}}
    & ResNet-50\cite{he2016deep} & 51.09 & 78.54 & 89.34 & 53.35 & 81.34 & 89.97 & 53.05 & 82.88 & 90.44 \\
    & ResNet-50+Occlusion Simulator & 55.48 & 80.71 & 90.54 & 58.42 & 83.11 & 91.28 & 59.96 & 83.59 & 91.08 \\
    & ResNet-50+Multi-task Losses & 54.97 & 82.11 & 90.54 & 58.88 & 84.41 & 91.28 & 59.76 & 84.66 & 91.08 \\
    & ResNet-50+Ours & \textbf{\textcolor{red}{58.15}} & \textbf{\textcolor{red}{84.61}} & \textbf{\textcolor{red}{92.11}} & \textbf{\textcolor{red}{61.31}} & \textbf{\textcolor{red}{85.94}} & \textbf{\textcolor{red}{92.84}} & \textbf{\textcolor{red}{63.55}} & \textbf{\textcolor{red}{87.02}} & \textbf{\textcolor{red}{93.50}} \\
    \hline

\end{tabular}
\end{table}

\begin{table}[!tp]
  \centering
  \caption{Detection precision of salient regions in saliency maps on Occluded-REID and Partial-REID} \label{tab:cap}
  \vspace{-0.2cm}
  \renewcommand\arraystretch{1} 
  \tabcolsep 14pt 
  \scriptsize  
  \doublerulesep=0.4pt
  \begin{tabularx}{\columnwidth}{|c|*{3}{Y|}}
    \hline
    METHODS & Occluded-REID & Partial-REID \\
    \hline
    ResNet-50\cite{he2016deep} & 77.26\% & 75.07\% \\
    ResNet-50+Ours & \textbf{\textcolor{red}{83.48\%}} & \textbf{\textcolor{red}{82.20\%}} \\
    \hline
  \end{tabularx}
\end{table}

\vspace{-0.3cm}
\begin{table}[!tp]
\begin{center}
  \caption{Comparison on rank-1/5/10 with state-of-the-art.} \label{tab:cap}
  \vspace{-0.2cm}
  \renewcommand\arraystretch{1.0} 
  \tabcolsep 0.5pt 
  \scriptsize  
  \doublerulesep=0.4pt

  \begin{tabular}{|p{0.1cm}|c|c|c|c|c|c|c|c|c|c|c|c|c|}
    \hline
  \multirow{2}{*}{\rotatebox{90}{\tiny Cat.}}
    & \multirow{1}{*}{\rotatebox{0}{DATA}} & \multicolumn{3}{c|}{Occluded-REID} & \multicolumn{3}{c|}{Partial-REID} & \multicolumn{3}{c|}{P-DukeMTMC} & \multicolumn{3}{c|}{P-ETHZ}\\
    \cline{2-14}
    & \multirow{1}{*}{\rotatebox{0}{METHODS}} & r=1 & r=5 & r=10 & r=1 & r=5 & r=10 & r=1 & r=5 & r=10 & r=1 & r=5 & r=10 \\
    \hline
  \multirow{4}{*}{\rotatebox{0}{\tiny A}}
    & XQDA\cite{liao2015person} & 36.71 & 65.11 & 77.65 & 33.14 & 66.18 & 83.03 & 15.93 & 27.50 & 33.89 & 44.98 & 70.88 & 83.69 \\
    & KCVDCA\cite{chen2015mirror} & 32.48 & 59.10 & 71.70 & 36.20 & 70.33 & 85.40 & 22.18 & 36.75 & 44.01 & 39.45 & 69.76 & 83.50 \\
    & GOG\cite{matsukawa2016hierarchical} & 40.50 & 63.16 & 73.77 & 41.92 & 74.00 & 86.54 & 17.10 & 29.27 & 35.82 & 49.17 & 79.29 & 90.21 \\
    & Null Space\cite{zhang2016learning} & 46.47 & 75.36 & 85.37 & 37.73 & 72.12 & 88.26 & 35.17 & 53.65 & 61.85 & 40.16 & 71.53 & 84.42 \\
  \hline  
\multirow{3}{*}{\rotatebox{0}{\tiny B}}
    & DGD\cite{xiao2016learning} & 41.43 & 65.74 & 75.91 & 56.83 & 77.70 & 85.00 & 41.53 & 60.09 & 67.61 & 51.23 & 81.01 & 91.16 \\
    & SVDNet\cite{sun2017svdnet} & 63.13 & 85.13 & 92.28 & 56.05 & 87.06 & 94.31 & 43.47 & 63.41 & \textbf{\textcolor{red}{71.12}} & 52.21 & 78.95 & 87.44 \\
    & REDA\cite{zhong2017random} & 65.79 & 87.88 & \textbf{\textcolor{red}{93.85}} & 76.19 & 94.57 & 97.43 & 45.18 & 62.88 & 69.92 & 54.43 & 79.09 & 89.17 \\
  \hline
  \multirow{2}{*}{\rotatebox{0}{\tiny C}}
    & ResNet-50\cite{he2016deep} & 57.27 & 81.40 & 89.66 & 69.82 & 89.68 & 96.10 & 39.16 & 57.13 & 64.35 & 51.09 & 78.54 & 89.34 \\
    & ResNet-50+Ours & \textbf{\textcolor{red}{68.14}} & \textbf{\textcolor{red}{88.29}} & 93.67 & \textbf{\textcolor{red}{78.52}} & \textbf{\textcolor{red}{94.87}} &  \textbf{\textcolor{red}{98.03}} & \textbf{\textcolor{red}{46.15}} & \textbf{\textcolor{red}{63.47}} & 70.67 & \textbf{\textcolor{red}{58.15}} & \textbf{\textcolor{red}{84.61}} & \textbf{\textcolor{red}{92.11}} \\
    
    \hline
  \end{tabular}
\end{center}
\vspace{-0.8cm}
\end{table}

\begin{figure*}[!t]
  \vspace{-0.8cm}
  \begin{center}
    \subfloat[Occluded-REID]{\label{fig:Per5}\includegraphics[width=0.25\textwidth]{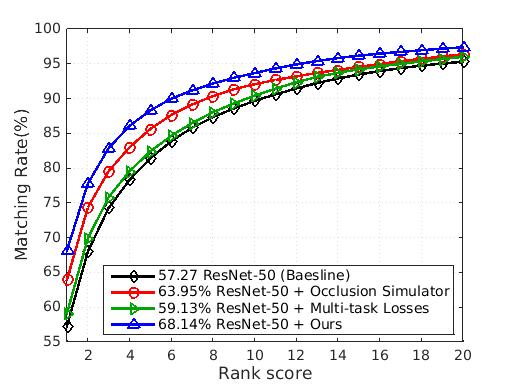}}
    \subfloat[Partial-REID]{\label{fig:Per4}\includegraphics[width=0.25\textwidth]{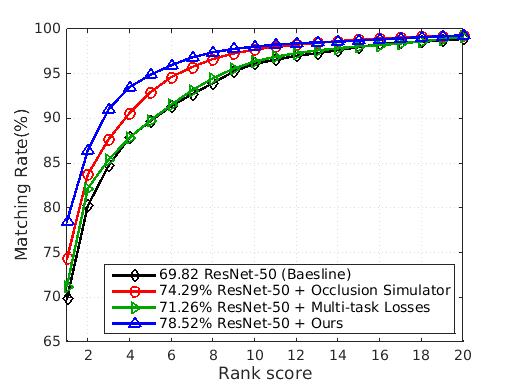}}
    \subfloat[P-DukeMTMC-reID]{\label{fig:Per5}\includegraphics[width=0.25\textwidth]{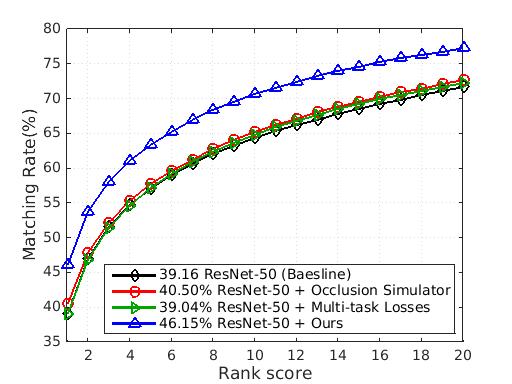}}
    \subfloat[P-ETHZ]{\label{fig:Per4}\includegraphics[width=0.25\textwidth]{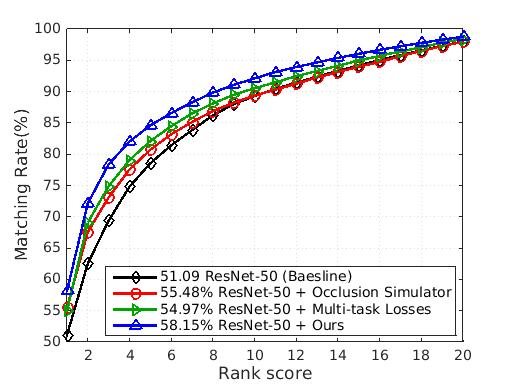}} 
    \vspace{-0.2cm}
    \caption{Comparisions with baseline on CMC curve and rank-1 rate.}
    \label{fig:oscil}
  \end{center}
  \vspace{-0.5cm}
\end{figure*}

\begin{figure*}[!t]
\vspace{-0.5cm}
  \begin{center}
    \subfloat[Occluded-REID]{\label{fig:Per5}\includegraphics[width=0.25\textwidth]{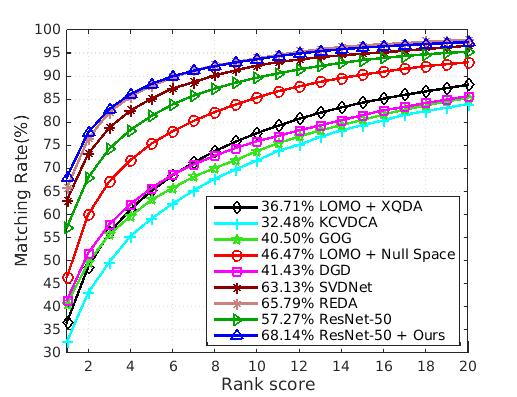}}
    \subfloat[Partial-REID]{\label{fig:Per4}\includegraphics[width=0.25\textwidth]{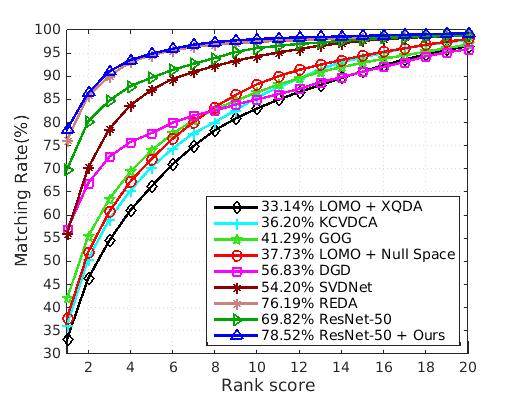}}
    \subfloat[P-DukeMTMC-reID]{\label{fig:Per5}\includegraphics[width=0.25\textwidth]{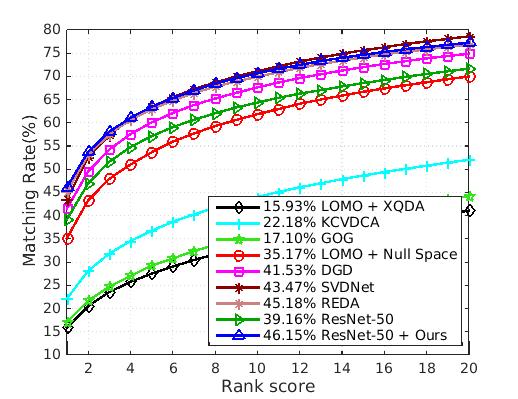}}
    \subfloat[P-ETHZ]{\label{fig:Per4}\includegraphics[width=0.25\textwidth]{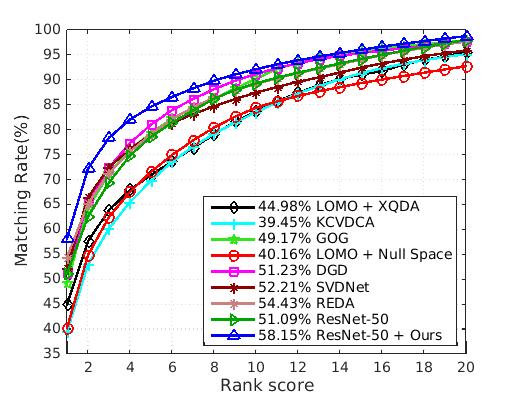}}
    \vspace{-0.2cm}
    \caption{Comparisions with state-of-the-art on CMC curve and rank-1 rate.}
    \label{fig:oscil}
  \end{center}
\vspace{-0.5cm}
\end{figure*}

\subsection{Analysis of parameter}

\begin{figure}[!t]
\vspace{-0.5cm}
  \begin{center}
    \subfloat[Occluded-REID]{\label{fig:Per5}\includegraphics[width=0.25\textwidth]{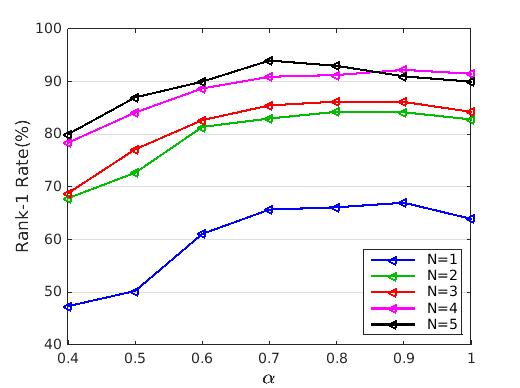}}
    \subfloat[Partial-REID]{\label{fig:Per4}\includegraphics[width=0.25\textwidth]{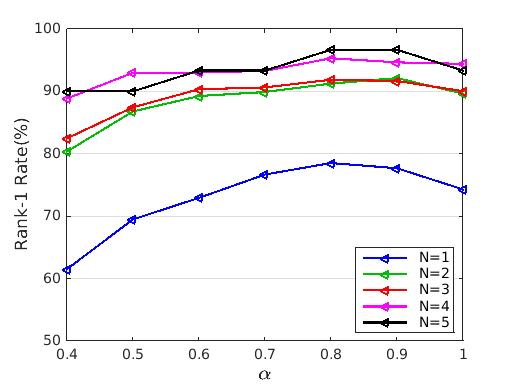}}
    \vspace{-0.2cm}
    \caption{Rank-1 rates of single-shot and multi-shot experiments on Occluded-REID and Partial-REID with different $\alpha$.}
    \label{fig:oscil}
  \end{center}
\vspace{-0.7cm}
\end{figure}

The only one parameter $\alpha$ in our model controls the tradeoff between the identification loss and OBC loss. To explore the effect of different proportion of two losses, we further test the performance of the representation with different $\alpha$ on Occluded-REID and Partial-REID. As is shown in Fig. 7, performance of the representation raises as the increase of $\alpha$. Our model achieves better performace when $\alpha$ is within 0.7 and 0.9, which confirms the auxiliary effect of the OBC loss.

\section{CONCLUSION}

In this paper, we make the first attempt to solve the occluded person re-id problem. To address it, the AFPB is proposed to learn a robust feature representation by watching kinds of generated occluded person images. Besides, multi-task losses are integrated into the framework for the attention of the person body. Experimental results show the effectiveness and superiority of our method.

This project is supported by the NSFC(U1611461, 61573387).

\let\oldthebibliography=\thebibliography
\let\endoldthebibliography=\endthebibliography
\renewenvironment{thebibliography}[1]{%
\begin{oldthebibliography}{#1}%
  \setlength{\parskip}{0ex}%
  \setlength{\itemsep}{0.1ex}%
  \footnotesize
}%
{%
\end{oldthebibliography}%
}
\begin{spacing}{0}
{
\bibliographystyle{IEEEbib}
\bibliography{icme2018}
}
\end{spacing}

\end{document}